\documentclass[11pt]{llncs}
\usepackage{setspace}
\usepackage{tocloft}
\usepackage{graphicx}
\usepackage{epstopdf}
\usepackage[bottom]{footmisc}
\usepackage[hidelinks]{hyperref}
\usepackage{color}
\usepackage{comment}
\usepackage{fixltx2e}
\usepackage{booktabs}
\usepackage{pgfplots}
\usepackage{pgfplotstable}
\usepackage{amsmath,amsfonts,amssymb}
\usepackage{algorithm}
\usepackage{enumitem}
\usepackage[noend]{algpseudocode}
\usepackage[numbers]{natbib}

\begin{document}

\title{Automatic Argumentative-Zoning Using Word2vec}


\author{Haixia Liu}
\institute{
School Of Computer Science, University of Nottingham Malaysia Campus, Jalan Broga, 43500 Semenyih, Selangor Darul Ehsan.\\
\email{khyx3lhi@nottingham.edu.my}\\ 
}

\maketitle

\begin{abstract}
In comparison with document summarization on the articles from social media and newswire, argumentative zoning (AZ) is an important task in scientific paper analysis. Traditional methodology to carry on this task relies on feature engineering from different levels. In this paper, three models of generating sentence vectors for the task of sentence classification were explored and compared. The proposed approach builds sentence representations using learned embeddings based on neural network. The learned word embeddings formed a feature space, to which the examined sentence is mapped to. Those features are input into the classifiers for supervised classification. Using 10-cross-validation scheme, evaluation was conducted on the Argumentative-Zoning (AZ) annotated articles. The results showed that simply averaging the word vectors in a sentence works better than the paragraph to vector algorithm and by integrating specific cuewords into the loss function of the neural network can improve the classification performance. In comparison with the hand-crafted features, the word2vec method won for most of the categories. However, the hand-crafted features showed their strength on classifying some of the categories. 
\end{abstract}

\section{Introduction}\label{sec:introduction}
One of the crucial tasks for researchers to carry out scientific investigations is to detect existing ideas that are related to their research topics. Research ideas are usually documented in scientific publications. Normally, there is one main idea stated in the abstract, explicitly presenting the aim of the paper. There are also other sub-ideas distributed across the entire paper. As the growth rate of scientific publication has been rising dramatically, researchers are overwhelmed by the explosive information. It is almost impossible to digest the ideas contained in the documents emerged everyday. Therefore, computer assisted technologies such as document summarization are expected to play a role in condensing information and providing readers with more relevant short texts. Unlike document summarization from news circles, where the task is to identify centroid sentences \cite{radev2000centroid} or to extract the first few sentences of the paragraphs \cite{lin1997identifying}, summarization of scientific articles involves extra text processing stage \cite{teufel2002summarizing}. After highest ranked texts are extracted, rhetorical status analysis will be conducted on the selected sentences. Rhetorical sentence classification, also known as argumentative zoning (AZ) \cite{teufel2000argumentative}, is a process of assigning rhetorical status to the extracted sentences. The results of AZ provide readers with general discourse context from which the scientific ideas could be better linked, compared and analyzed. For example, given a specific task, which sentences should be shown to the reader is related to the features of the sentences. For the task of identifying a paper's unique contribution, sentences expressing research purpose should be retrieved with higher priority. For comparing ideas, statements of comparison with other works would be more useful. Teufel et. al. \cite{teufel2002summarizing} introduced their rhetorical annotation scheme which takes into account of the aspects of argumentation, metadiscourse and relatedness to other works. Their scheme resulted seven categories of rhetorical status and the categories are assigned to full sentences. Examples \footnote{These texts were randomly selected from Argumentative Zoning Corpus, which is described in dataset section.} of human annotated sentences with their rhetorical status are shown in Table.~\ref{table:examples}. The seven categories are aim, contrast, own, background, other, basis and textual.  

\begin{table*}[!htbp]
\caption{Examples of annotated sentences with their rhetorical status }
\scriptsize
\centering   
\begin{tabular}{c c}   
\hline\hline  
 Rhetorical Status &  Examples  \\ [0.5ex]
\hline 
 AIM & \multicolumn{1}{l}{This paper discusses the lexicographical concept of lexical functions}\\
     & \multicolumn{1}{l}{Mel'cuk and Zolkovsky 1984 and their potential exploitation in the} \\
     & \multicolumn{1}{l}{development of a machine translation lexicon designed to handle collocations.} \\
     \hline 
 CTR & \multicolumn{1}{l}{In two of the tasks, the training data is generated by} \\
     & \multicolumn{1}{l}{a probabilistic context-free grammar and in both tasks our}\\
     & \multicolumn{1}{l}{algorithm outperforms the other techniques.}\\
     \hline 
 OWN & \multicolumn{1}{l}{We have explored examples of the kinds of tree sets} \\
     & \multicolumn{1}{l}{and string languages that this system can generate.}\\
     \hline
 BKG & \multicolumn{1}{l}{English has a very limited system, marking little} \\
     & \multicolumn{1}{l}{more than plurality on nouns and a restricted range of verb properties.}\\
     \hline 
 OTH & \multicolumn{1}{l}{For this small example, writing such an apply predicate} \\
     & \multicolumn{1}{l}{is not difficult.}\\
     \hline
 BAS & \multicolumn{1}{l}{Following Pereira et al. 1993, we measure word} \\
     & \multicolumn{1}{l}{similarity by the relative entropy, or Kullback-Leibler} \\
     & \multicolumn{1}{l}{distance, between the corresponding conditional distributions.} \\
     \hline 
 TXT & \multicolumn{1}{l}{The next section describes the binary representation} \\
     & \multicolumn{1}{l}{and the length formul derived from it in detail; } \\
     & \multicolumn{1}{l}{readers satisfied with the intuitive descriptions} \\
     & \multicolumn{1}{l}{presented so far should skip ahead to the Phonotactics sub-section.}\\
     
\hline             
\end{tabular}
\label{table:examples} 
\end{table*}

Analyzing the rhetorical status of sentences manually requires huge amount of efforts, especially for structuring information from multiple documents. Fortunately, computer algorithms have been introduced to solve this problem. With the development of artificial intelligence, machine learning and computational linguistics, Natural Language Processing (NLP) has become a popular research area \cite{manning1999foundations,hirschberg2015advances}. NLP covers the applications from document retrieval, text categorization \cite{jackson2007natural}, document summarization \cite{cao2015ranking} to sentiment analysis \cite{nasukawa2003sentiment,pang2008opinion}. Those applications are targeting different types of text resources, such as articles from social media \cite{asur2010predicting} and scientific publications \cite{teufel2002summarizing}. There are several approaches to tackle these tasks. From machine learning prospective, text can be analysed via supervised \cite{teufel2002summarizing}, semi-supervised \cite{sindhwani2008document} and unsupervised \cite{hofmann2001unsupervised} algorithms.   

Document summarization from social media and news circles has received much attention for the past decades. Those problems have been addressed from many angles, one of which is feature extraction and representation. At the early stage of document summarization, features are usually engineered manually. Although the hand-crafted features have shown the ability for document summarization and sentiment analysis \cite{das2007survey,pang2008opinion}, there are not enough efficient features to capture the semantic relations between words, phrases and sentences. Moreover, building a sufficient pool of features manually is difficult, because it requires expert knowledge and it is time-consuming. Teufel et. al. \cite{teufel2002summarizing} have built feature pool of sixteen types of features to classify sentences, such as the position of sentence, sentence length and tense. Widyantoro et. al. used content features, qualifying adjectives and meta-discourse features \cite{widyantoro2015multiclass} to explore AZ task. It took efforts to engineer these features and it is also time consuming to optimize the combination of the entire features.  
With the advent of neural networks \cite{hinton2006fast}, it is possible for computers to learn feature representations automatically. Recently, word embedding technique \cite{mikolov2013efficient} has been widely used in the NLP community. There are plenty of cases where word embedding and sentence representations have been applied to short text classification \cite{wang2015semantic} and paraphrase detection \cite{socher2011dynamic}. However, the effectiveness of this technique on AZ needs further study. The research question is, is it possible to extract word embeddings as features to classify sentences into the seven categories mentioned above using supervised machine learning approach? 
\section{Related Work}\label{sec:related_work}
The tool of word2vec proposed by Mikolov et al. \cite{mikolov2013efficient} has gained a lot attention recently. With word2vec tool, word embeddings can be learnt from big amount of text corpus and the semantic relationships between words can be measured by the cosine distances between the vectors. The idea behind word embeddings is to use distributed representation \cite{Hinton86} to map each word into k-dimension vector. How these vectors are generated using word2vec tool? The common method to derive the vectors is using neural probabilistic language model \cite{bengio2003neural}. The underlying word representations for each word are obtained while training the language model. Similar to the mechanism in language model, Mikolov et al. \cite{mikolov2013efficient} introduced two architectures: Skip-gram model and continuous bag of words (CBOW) model. Each of the model has two different training strategies, such as hierarchical softmax and negative sampling. Both these two models have three layers: input, projection and output layer. The word vectors are obtained once the models are optimized. Usually, this optimizing process is done using stochastic gradient descent method. It doesn't need labels when training the models, which makes word2vec algorithm more valuable compared with traditional supervised machine learning methods that require a big amount of annotated data. Given enough text corpus, the word2vec can generate meaningful representations.

Word2vec has been applied to sentiment analysis \cite{tang2014learning,xue2014study,zhang2015chinese} and text classification \cite{lilleberg2015support}. Sadeghian and Sharafat \cite{sadeghianbag} explored averaging of the word vectors in a sentiment review statement. Their results indicated that word2vec models significantly outperform the vanilla bag-of-words model. Amongst the word2vec based models, softmax provides the best form of classification. Tang et al. \cite{tang2014learning} used the concatenation of vectors derived from different convolutional layers to analyze the sentiment statements. They also trained sentiment-specific word embeddings to improve the twitter sentiment classification results.  This work is aiming at learning word embeddings for the task of AZ. The results were compared from three aspects: the impact of the training corpus, the effectiveness of specific word embeddings and different ways of constructing sentence representations based on the learned word vectors.

Le and Mikolov \cite{le2014distributed} introduced the concept of word vector representation in a formal way:

Given a sequence of training words $w = <w_{1},x_{2},...,w_{n}>$, the objective of the word2vec model is to maximize the average log probability: \\

$\frac{1}{T}$ $\sum^{T-k}_{t=k}$ $log$ p$(w_{t}|w_{t-k},...,w_{t+k})$ (1) \\

Using softmax technique, the prediction can be formalized as:\\

p$(w_{t}|w_{t-k},...,w_{t+k})$ = $\frac{e^{y_{w_{t}}}}{\sum e^{y_{w_{y}}}}$ (2) \\

Each of $y_{i}$ is un-normalized log probability for each output word $i$: \\

$y =  b + Uh(w_{t-k},...,w_{t+k};W)$ (3) \\
\section{Methodology}\label{sec:Methodology}
\subsection{Models}\label{sec:models}
In this study, sentence embeddings were learned from large text corpus as features to classify sentences into seven categories in the task of AZ. Three models were explored to obtain the sentence vectors: averaging the vectors of the words in one sentence, paragraph vectors and specific word vectors.  

The first model, averaging word vectors ($AVGWVEC$), is to average the vectors in word sequence $w = <w_{1},x_{2},...w_{n}>$. The main process in this model is to learn the word embedding matrix $W_{w}$:\\

{$V_{avgwvec}(w) = $ $\frac{1}{n}$ $\sum$ $W^{x_{i}}_{w}$} (4)\\

where $W_{w}$ is the word embedding for word $x_{i}$, which is learned by the classical word2vec algorithm \cite{mikolov2013efficient}. 

The second model, $PARAVEC$, is aiming at training paragraph vectors. It is also called distributed memory model of paragraph vectors (PV-DM) \cite{le2014distributed}, which is an extension of word2vec. In comparison with the word2vec framework, the only change in PV-DM is in the equation (3), where $h$ is constructed from $W$ and $D$, where matrix $W$ is the word vector and $D$ holds the paragraph vectors in such a way that every paragraph is mapped to a unique vector represented by a column in matrix $D$.


The third model is constructed for the purpose of improving classification results for a certain category. In this study specifically, the optimization task was focused on identifying the category $BAS$ \footnote{This is a general case to show how to improve the classification result by integrating cuewords to the embeddings.}. In this study, $BAS$ specific word embeddings were trained ($BSWE$) inspired by Tang et al. \cite{tang2014learning}'s model: Sentiment-Specific Word Embedding (unified model: $SSWE_{u}$). After obtaining the word vectors via $BSWE$, the same scheme was used to average the vectors in one sentence as in the model $AVGWVEC$.

\subsection{Classification and evaluation}\label{sec:classification_eval}
The learned word embeddings are input into a classifier as features under a supervised machine learning framework. Similar to sentiment classification using word embeddings \cite{tang2014learning}, where they try to predict each tweet to be either positive or negative, in the task of AZ, the embeddings are used to classify each sentence into one of the seven categories.

To evaluate the classification performance, precision, recall and F-measure were computed. 

\section{Experimental Evaluation}
\subsection{Training Dataset}
$ACL$ collection. ACL Anthology Reference Corpus \footnote{$http://acl$-$arc.comp.nus.edu.sg/$} contains the canonical 10,921 computational linguistics papers, from which 622,144 sentences were generated after filtering out sentences with lower quality. 
 
$MixedAbs$ collection contains 6,778 sentences, extracted from the titles and abstracts of publications provided by WEB OF SCIENCE \footnote{$webofknowledge.com$}.

\subsection{Test Dataset}
Argumentative Zoning Corpus ($AZ$ corpus) consists of 80 AZ$-$annotated conference articles in computational linguistics, originally drawn from the Cmplg arXiv. \footnote{$http://www.cl.cam.ac.uk/$\textasciitilde$sht25/AZ\_corpus.html$}. After Concatenating sub-sentences, 7,347 labeled sentences were obtained. 

\subsection{Training strategy}
To compare the three models effectiveness on the AZ task, the three models on a same ACL dataset (introduced int he dataset section) were trained. The word2vec were also trained using different parameters, such as different dimension of features. To evaluate the impact from different domains, the first model was trained on different corpus.

The characteristics of word embeddings based on different model and dataset are listed in Table.~\ref{table:characterwe}.

\begin{table*}[!htbp]
\caption{Characteristics of word embeddings based on different model and dataset }
\scriptsize
\centering   
\begin{tabular}{c c c}   
\hline\hline  
 & Number of features  & Vocabulary size \\ [0.5ex]
\hline 
 $AVGWVEC$ ACL+AZ 300 & 300 & 13685 \\
 $AVGWVEC$ ACL+AZ 100 & 100 & 14325 \\
 $PARAVEC$ ACL+AZ 100 & 100 & 74261 \\
 $AVGWVEC$ MixedAbs 100 & 100 & 4126 \\
 $AVGWVEC$ $BSWE$ 100 & 100 & 643 \\
 $AVGWVEC$ Brown model & 100 & 56057\\
\hline             
\end{tabular}
\label{table:characterwe} 
\end{table*}

\subsection{Parameters}
Inspired by the work from Sadeghian and Sharafat \cite{sadeghianbag} \footnote{$https://www.kaggle.com/c/word2vec-nlp-tutorial/details/part-2-word-vectors$}, the word to vector features were set up  as follows: the Minimum word count is 40; The number of threads to run in parallel is 4 and the context window is 10. 

\subsection{Strategy of dealing with unbalanced data}
In imbalanced data sets, some classes are significantly outnumbered by other classes \cite{nguyen2011borderline}, which affects the classification results. In this experiment, the test dataset is an imbalanced data set. Table. ~\ref{table:districat} shows the distribution of rhetorical categories from the $AZ$ test dataset. The categories OWN and OTH are significantly outnumbering other categories.

\begin{table*}[!htbp]
\caption{Distribution of rhetorical categories}
\scriptsize
\centering   
\begin{tabular}{c c c}   
\hline\hline  
 Category & Number of Sentences & Percentage \\ [0.5ex]
\hline 
OWN & 4868 & 0.54\\
OTH & 1927 & 0.21\\
BKG & 644 & 0.07\\
BAS & 641 & 0.07\\
CTR & 451 & 0.05\\
AIM & 303 & 0.03\\
TXT & 191 & 0.02\\
\hline             
\end{tabular}
\label{table:districat} 
\end{table*}

To deal with the problem of classification on unbalanced data, synthetic Minority Over-sampling TEchnique (SMOTE) \cite{chawla2002smote} were performed on the original dataset. 10-cross validation scheme was adopted and the results were averaged from 10 iterations. 

\subsection{Results of classification for per category}
Table.~\ref{table:results1} and ~\ref{table:results2} show the classification performance of different methods.
\footnote{Note that it is not completely compatible with Teufel 2002 results, since the dataset is different due to the sentence concatenation in this paper. But Teufel's reports could be a reference.} 

\begin{table*}[!htbp]
\caption{Performance of sentence classification per category I (precision/recall/F-measure) }
\scriptsize
\centering   
\begin{tabular}{c c c c c }   
\hline\hline  
Method & AIM & CTR & BKG & BAS \\ [0.5ex]
\hline 
$AVGWVEC$ ACL+AZ 300 & 0.29/0.82/0.43 & 0.34/0.75/0.47 & 0.36/0.72/0.48 & 0.10/0.72/0.17 \\
$AVGWVEC$ ACL+AZ 100 & 0.29/0.85/0.43 & 0.29/0.80/0.42 & 0.36/0.68/0.47 & 0.11/0.87/0.20 \\
$PARAVEC$ ACL+AZ 100 & 0.60/0.03/0.06 & 0.20/0.004/0.009 &  0.39/0.02/0.04 & 0.00/0.00/0.00\\
$AVGWVEC$ MixedAbs 100 & 0.11/0.73/0.19 & 0.11/0.71/0.20 & 0.14/0.62/0.23 & 0.04/0.65/0.08 \\
$AVGWVEC$ Brown model 100 & 0.19/0.73/0.30 & 0.38/0.56/0.45 & 0.19/0.55/0.28 & 0.05/0.72/0.10 \\
$AVGWVEC$ $BSWE$ 100 & - & - & - & 0.14/0.63/0.23 \\ 
Cuewords & 0.13/0.55/0.21 & 0.33/0.20/0.25 & - & 0.08/0.36/0.13 \\
Teufel 2002 & 0.44/0.65/0.52 & 0.34/0.20/0.26 & 0.40/0.50/0.45 & 0.37/0.40/0.38 \\
Baseline & 0.30/0.07/0.11 & 0.31/0.12/0.17 & 0.32/0.17/0.22 & 0.15/0.05/0.07 \\
\hline             
\end{tabular}
\label{table:results1} 
\end{table*}

\begin{table*}[!htbp]
\caption{Performance of sentence classification per category II (precision/recall/F-measure)}
\scriptsize
\centering   
\begin{tabular}{c c c c}   
\hline\hline  
 & TXT & OWN & OTH \\ [0.5ex]
\hline 
$AVGWVEC$ ACL+AZ 300 & 0.51/0.87/0.64 & 0.61/0.71/0.65 & 0.49/0.65/0.56 \\
$AVGWVEC$ ACL+AZ 100 & 0.47/0.88/0.61 & 0.59/0.68/0.63 & 0.49/0.69/0.57 \\
$PARAVEC$ ACL+AZ 100 & 0.52/0.11/0.18 & 0.62/0.98/0.76 & 0.35/0.004/0.009 \\ 
$AVGWVEC$ MixedAbs 100 & 0.15/0.75/0.25 & 0.72/0.56/0.63 & 0.21/0.61/0.31  \\
$AVGWVEC$ Brown model 100 & 0.30/0.72/0.42 & 0.56/0.52/0.54 & 0.42/0.66/0.51 \\ 

Teufel 2002 & 0.57/0.66/0.61 & 0.84/0.88/0.86 & 0.52/0.39/0.44 \\
Baseline & 0.56/0.15/0.23 & 0.78/0.90/0.83 & 0.47/0.42/0.44 \\
\hline             
\end{tabular}
\label{table:results2} 
\end{table*}

The results were examined from the following aspects: 

When the feature dimension is set to 100 and the training corpus is ACL, the results generated by different models were compared (AVGWVEC,\\
PARAVEC and AVGWVEC+BSWE for BAS category only). Looking at the F-measure, AVGWVEC performs better than PARAVEC, but PARAVEC gave a better precision results on several categories, such as AIM, CTR, TXT and OWN. The results showed that PARAVEC model is not robust, for example, it performs badly for the category of BAS. For specific category classification, take the BAS category for example, the BSWE model outperforms others in terms of F-measure.   

When the model is fixed to AVGWVEC and the training corpus is ACL, the feature size impact (300 and 100 dimensions) was  investigated. From the F-measure, it can be seen that for some categories, 300-dimension features perform better than the 100-dimension ones, for example, CTR and BKG, but they are not as good as 100-dimension features for some categories, such as BAS.

When the model is set to AVGWVEC and the feature dimension is 100, the results computed from different training corpus were compared (ACL+AZ, MixedAbs and Brown corpus). ACL+AZ outperforms others and brown corpus is better than MixedAbs for most of the categories, but brown corpus is not as good as MixedAbs for the category of OWN.

Finally, the results  were compared between word embeddings and the methods of cuewords, Teufel 2002 and baseline. To evaluate word embeddings on AZ, the model AVGWVEC trained on ACL+AZ was used for the comparison. It can be seen from the table.~\ref{table:results1}, the model of word embeddings is better than the method using cuewords matching. It also outperforms Teufel 2002 for most of the cases, except AIM, BAS and OWN. It won baseline for most of the categories, except OWN.  
\section{Discussion}
The classification results showed that the type of word embeddings and the training corpus affect the AZ performance. As the simple model, $AVGWVEC$ performs better than others, which indicate averaging the word vectors in a sentence can capture the semantic property of statements. By training specific argumentation word embeddings, the performance can be improved, which can be seen from the case of detecting BAS status using $BSWE$ model.  

Feature dimension doesn't dominate the results. There is no significant difference between the resutls generated by 300-dimension of features and 100 dimensions.

Training corpus affects the results. ACL+AZ outperforming others indicates that the topics of the training corpus are important factors in argumentative zoning. Although Brown corpus has more vocabularies, it doesn't win ACL+AZ. 

In general, the classification performance of word embeddings is competitive in terms of F-measure for most of the categories. But for classifying the categories AIM, BAS and OWN, the manually crafted features proposed by Teufel et al. \cite{teufel2002summarizing} gave better results.

\section{Conclusion}
In this paper, different word embedding models on the task of argumentative zoning were compared . The results showed that word embeddings are effective on sentence classification from scientific papers. Word embeddings trained on a relevant corpus can capture the semantic features of statements and they are easier to be obtained than hand engineered features.

To improve the sentence classification for a specific category, integrating word specific embedding strategy helps. The size of the feature pool doesn't matter too much on the results, nor does the vocabulary size. In comparison, the domain of the training corpus affects the classification performance.


\end{document}